%% file: root.tex
\documentclass[letterpaper, 10 pt, conference]{ieeeconf}  

\IEEEoverridecommandlockouts                              

\usepackage{amsmath} 
\usepackage{bm}
\usepackage{xcolor}
\usepackage{diagbox}
\usepackage{multirow}
\usepackage{multicol}
\usepackage{graphicx}
\usepackage{soul}
\usepackage{wrapfig}
\usepackage{hyperref}
\usepackage{ulem}

\title{\LARGE \bf
Multi-finger Manipulation via Trajectory Optimization with Differentiable Rolling and Geometric Constraints
}

\author{Fan Yang$^1$, Thomas Power$^1$, Sergio Aguilera Marinovic$^2$, Soshi Iba$^2$, Rana Soltani Zarrin$^2$, Dmitry Berenson$^1$
\thanks{$^{1}$Robotics Department, 
        University of Michigan, Ann Arbor, MI, USA
        {\tt\small [fanyangr,tpower,dmitryb]@umich.edu}  $^{2}$ Honda Research Institute USA. This work was sponsored by Honda Research Institute USA.}%
}

\begin{document}

\maketitle
\thispagestyle{empty}
\pagestyle{empty}

\begin{abstract}
Parameterizing finger rolling and finger-object contacts in a differentiable manner is important for formulating dexterous manipulation as a trajectory optimization problem. In contrast to previous methods which often assume simplified geometries of the robot and object or do not explicitly model finger rolling, we propose a method to further extend the capabilities of dexterous manipulation by accounting for non-trivial geometries of both the robot and the object. By integrating the object's Signed Distance Field (SDF) with a sampling method, our method estimates contact and rolling-related variables in a differentiable manner and includes those in a trajectory optimization framework. This formulation naturally allows for the emergence of finger-rolling behaviors, enabling the robot to locally adjust the contact points. To evaluate our method, we introduce a benchmark featuring challenging multi-finger dexterous manipulation tasks, such as screwdriver turning and in-hand reorientation. Our method outperforms baselines in terms of achieving desired object configurations and avoiding dropping the object. We also successfully apply our method to a real-world screwdriver turning task and a cuboid alignment task, demonstrating its robustness to the sim2real gap. 
\end{abstract}

\section{INTRODUCTION}
\footnote{This work has been submitted to the IEEE for possible publication. Copyright may be transferred without notice, after which this version may no longer be accessible.}
\input{intro}
\section{RELATED WORK}
\input{related_works}
\section{PROBLEM STATEMENT}
\input{problem_statement}
\section{METHOD}
\input{method}

\section{EXPERIMENTS}

\input{experiments}
\vspace{-0.2cm}
\section{CONCLUSION}
\vspace{-0.2cm}
\input{discussion}
\vspace{-0.3cm}
\bibliographystyle{IEEEtran}
\bibliography{reference}

\end{document}

%% file: intro.tex
Multi-finger dexterous manipulation can be used to accomplish a wide range of useful dexterous manipulation tasks, such as turning a screwdriver or orienting an object. Many methods have approached such tasks via reinforcement learning (RL)~\cite{chen2022system, openai_dexterous, ma2023eureka}. While these methods are capable of producing the desired behavior given extensive task-specific training, they provide no guarantees on constraint satisfaction (e.g. maintaining finger contact). They are also ill-suited for dynamic task assignment, as a large amount of data must be collected for each new task.

In this paper, we focus on the problem of dexterous manipulation with a fixed set of contact modes using a multi-fingered hand, i.e. where fingers do not make new contacts or break contact during the task. A method that addresses this manipulation problem can be used with a higher-level planner or learning algorithm that outputs a sequence of contact modes to accomplish multi-stage tasks (e.g. finger-gaiting)~\cite{xue2023dynamic, zarrin2023hybrid, morgan2022complex}. We approach this problem using trajectory optimization. While this approach requires more modeling of the system than an RL method, it is more extensible to new problems and does not require training data, so it is appropriate for a setting where new tasks may arise dynamically.
\begin{figure}
    \centering
    \includegraphics[width=0.8\linewidth]{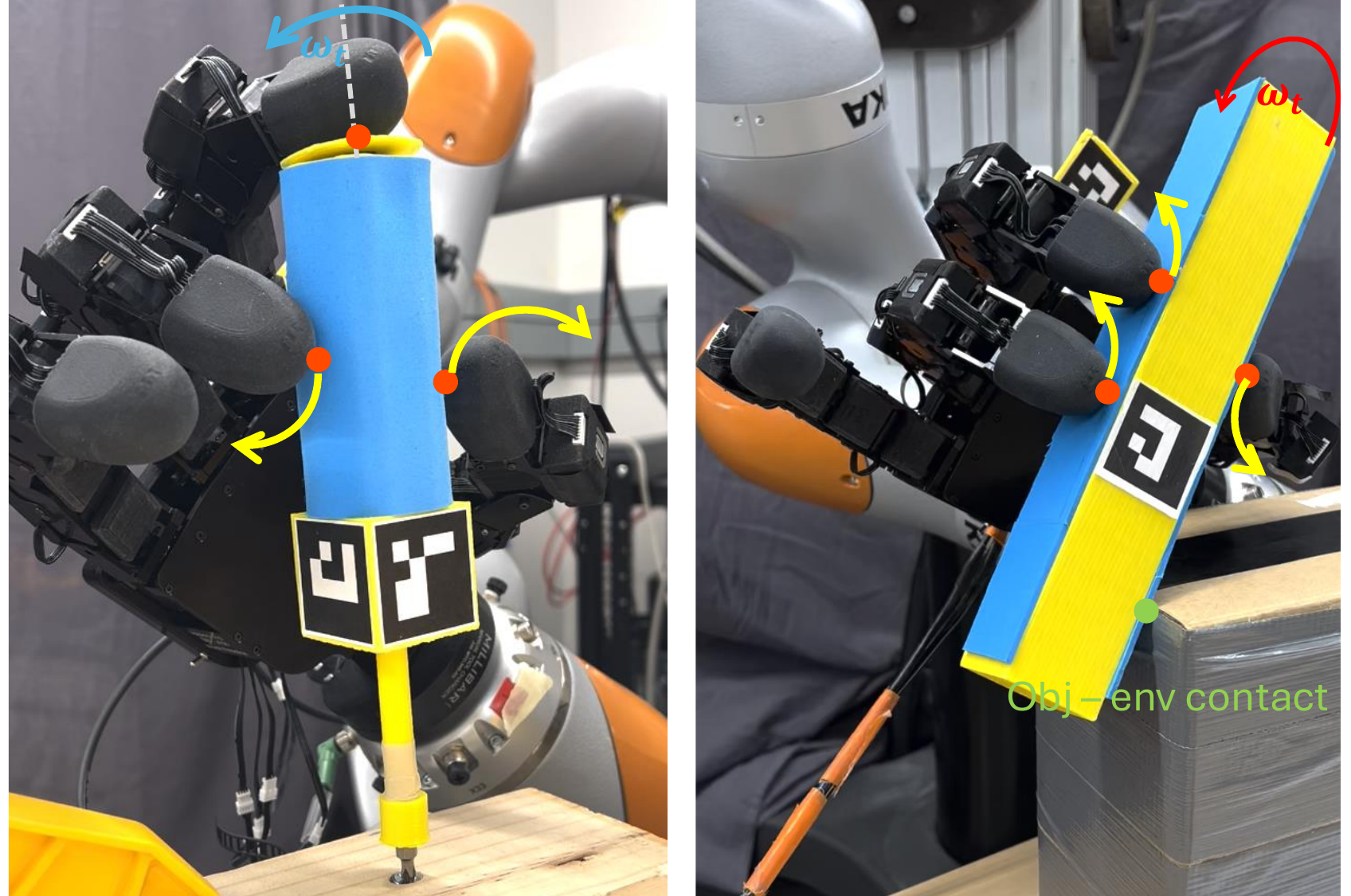}
    \vspace{-0.2cm}
    \caption{Rolling the finger can extend the flexibility of the robot and is necessary for many dexterous manipulation tasks. Parameterizing the geometry of both the robot fingers and the manipulated object is important for formulating the finger-object contacts and rolling behaviors. We propose a sampling method to approximate the geometry of the robots. Integrating it with the SDF of the object, contact-related variables such as distance between meshes, contact points, and contact normals can be estimated in a differentiable manner. 
    }
    \label{fig:pull_figure}
\end{figure}
There are two types of trajectory optimization that could be applied to this problem: sampling-based~\cite{williams2017information, rubinstein1999cross} and gradient-based~\cite{power2024constrained, ratliff2009chomp}. While the sampling-based methods are more flexible, as they do not require differentiable dynamics or cost functions, they do not perform well under stringent constraints (e.g. maintaining finger contact). Thus we choose a gradient-based method: Constrained Stein Variational Trajectory Optimization(CSVTO)~\cite{power2024constrained}. This method decouples the cost and constraint gradients, ensuring constraint satisfaction even in low-dimensional manifolds like those induced by contact.

However, a key challenge for the gradient-based approach is formulating the constraints of the dexterous manipulation problem so that they are 1) differentiable; 2) account for non-trivial geometries in 3D; and 3) enable finger rolling to allow sufficient freedom in the contact interaction to perform useful tasks. Finally, we wish for the approach to be as reactive as possible, to account for unexpected changes in object state and model inaccuracies, and thus we will use our method inside a Model Predictive Control (MPC) framework. Consequently, the optimization should be formulated in a way that is computationally tractable, i.e. yielding acceptable results in a relatively small number of iterations.

To address these challenges, we make 2 key contributions:
\begin{enumerate}
 \item We formulate differentiable constraints for 3D finger-rolling and finger-object contacts, which consider the non-trivial geometry of robot fingers. This differentiable formulation enables trajectory optimization for sensitive manipulation tasks and allows significant freedom in contact interaction, enabling the method to use finger rolling to accomplish large object orientation changes without breaking contact.

 \item We propose a benchmark for multi-finger manipulation tasks, requiring the robot to adhere to stringent contact constraints to reach the goal and prevent irrecoverable failures, such as dropping the manipulated object.
\end{enumerate}
 
We evaluate our method on our benchmark which features object reorientation tasks both with and without external environment-object contacts.
Our simulated and real-world results suggest that formulating differentiable contact and finger-rolling constraints can significantly extend the capabilities of dexterous manipulation. 
Please see \url{https://sites.google.com/umich.edu/multi-finger-rolling/home} for more details about experiments, the appendix, and the benchmark.

%% file: related_works.tex
\subsection{Planning for Dexterous Manipulation}
There is a long history of developing planning methods for dexterous manipulation \cite{131782, fingergaiting, doi:10.1177/02783640122067480, Chen2021TrajectoTreeTO}. Many of these methods explicitly reason about changes in contact state. For instance, Xu et al. \cite{fingergaiting} propose sample-based motion planning in a hybrid configuration space for finger gait planning. Cheng et al. propose an algorithm~\cite{cheng2023enhancing}, which automatically generates contact modes in a sample-based framework. 
Optimization-based approaches have also been used for dexterous manipulation \cite{10.1145/1531326.1531365, sundaralingam2019relaxed, rozzi2024combining, mason2001mechanics, posa_contact_implicit, schulman2014motion}.
However, these methods do not explicitly consider the non-trivial geometries of the robot fingers, i.e., they approximate the robot fingertips using either a point or basic primitives such as a sphere and lack explicit differentiable modeling of finger contact or rolling behavior. Pang et al.~\cite{pang2023global} utilizes Drake~\cite{drake} to compute gradients of the contact and rolling constraints. 
However, their implementation simplifies all geometries to basic primitives to address the discontinuity of contact-related gradients.
To the best of our knowledge, this work is the first to incorporate non-trivial robot finger geometry in multi-finger dexterous manipulation tasks.


\subsection{Finger Rolling}
There have been many works exploiting rolling contact for dexterous manipulation \cite{rolling_cole, rolling_ball_plane, rolling_ball_plane_2, rolling_finger_gaiting, nonprehensile_rolling, planning_rolling_sliding}. Early methods were limited to the case of a spherical object rolling on a planar surface \cite{rolling_ball_plane, rolling_ball_plane_2}, or with semi-spherical fingertips \cite{rolling_finger_gaiting}. Bai and Liu \cite{nonprehensile_rolling} proposed a method that allows for rolling contacts with more general polyhedral geometries but it is limited to non-prehensile tasks. Recent work by Tang et al. \cite{tangscrewdriver} proposed a method for turning a screwdriver with a dexterous hand, that also exploits rolling contacts, however, this approach is highly specialized to the task. In contrast, our proposed method allows rolling contact with arbitrary finger geometries and object geometries consisting of a composition of primitives. We also show our method performing five distinct tasks.

\subsection{Benchmarks for Multi-finger Dexterous Manipulation}
Existing benchmarks for multi-finger manipulation include Chen et al.\cite{chen2022towards}, which focuses on human-like bimanual manipulation; Bao et al.\cite{bao2023dexart}, which emphasizes articulated objects; and Cruciani et al.~\cite{cruciani2020benchmarking}, which solely targets in-hand reorientation. However, no benchmark addresses fine multi-finger manipulation requiring precise control to prevent irrecoverable failures, such as dropping the object.

%% file: problem_statement.tex
In this work, we address dexterous manipulation with a multi-fingered hand, modeled as a discrete-time trajectory optimization problem, with the state $\mathbf{s}_t$, and the action $\mathbf{u}_t$. The state $\mathbf{s}_t$ includes the robot configuration $\mathbf{q}_t$, and object configuration: $\mathbf{o}_t:=\{\bm{\theta}_t ,\mathbf{x}_t\}$, which consists of object position $\mathbf{x}_t$ and object orientation $\bm{\theta}_t$. 
We assume the existence of a low-level controller (i.e. a PD controller) responsible for executing position commands and define the action space as the delta actions, which means $\mathbf{q}_t + \mathbf{u}_t$ is the commanded next robot configuration. 
The objective is to reach a desired object configuration $\mathbf{o_g}$. We model our problem in a trajectory optimization framework, aiming to optimize for a trajectory $\mathbf{\tau} := \{\mathbf{s}_1, \mathbf{u}_1, \cdots, \mathbf{s}_T, \mathbf{u}_T\}$, such that the objective $J(\mathbf{\tau})$ is minimized.

However, solving a general dexterous manipulation problem without additional assumptions is extremely challenging due to the high dimensionality of the trajectory and the non-differentiable nature of contact dynamics.
To make the problem tractable, we assume: (1) the system is quasi-static, (2) the contact modes are prespecified, and (3) the fingers are already in contact with the object initially. We further assume (4) knowledge of the geometry of robot fingers.

\begin{figure}
    \centering
    \vspace{0.3cm}
    \includegraphics[width=1.0\linewidth]{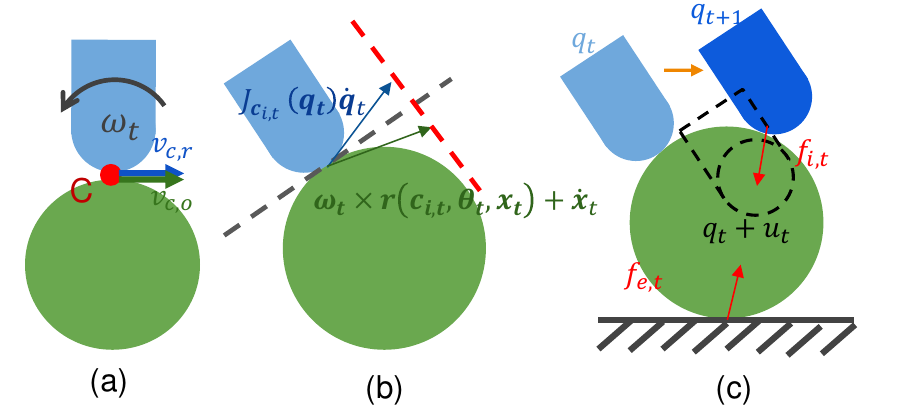}
    \vspace{-0.6cm}
    \caption{(a) Pure Rolling. The blue(robot) and the green(object) rigid bodies contact at the red point $\mathbf{c}$. The contact points have velocities $\mathbf{v}_{c,r}$ and $\mathbf{v}_{c,o}$ respectively. Pure rolling happens when the contact point pair has the same velocity: $\mathbf{v}_{c,r} = \mathbf{v}_{c,o}$. (b) Projection of the kinematics constraint into the tangential space of the contact. Specifically, the constraint is satisfied if the blue and the green vector both lie on the red dashed line. (c) Variables used in the force balance. The robot wants to move from $\mathbf{q}_t$ to $\mathbf{q}_t + \hat{\mathbf{u}}_t$ but ends up at $\mathbf{q}_{t+1}$. Thus there will be an end effector force $\mathbf{f}_{i,t}$ approximately pointing in the $\mathbf{q}_t + \hat{\mathbf{u}}_t$ direction.
    }
    \label{fig:rolling_and_kinematics}
\end{figure}

The contact mode is defined as the type of interaction between two objects~\cite{balkcom2002computing}, e.g., sliding, rolling, and not in contact. Pure rolling occurs when the contact point velocities of the robot and object are equal ( $\mathbf{v}_{c,r} = \mathbf{v}_{c,o}$, Fig.~\ref{fig:rolling_and_kinematics}(a))
Sliding occurs when $\mathbf{v}_{c,r} \neq \mathbf{v}_{c,o}$. Contact points are not created or removed within a contact mode. Specifically, in our problem, we assume a three-finger rolling contact and the robot hand initially grasps the object with its fingertips. It also maintains this grasp throughout the manipulation. 

Despite having a fixed contact mode, our method still gives the robot the flexibility of adjusting the contacts by rolling its fingertips. It is not trivial to model finger rolling since modeling arbitrary finger and object shapes and extracting meaningful gradients based on the geometries can be very challenging.

We will evaluate our method primarily in terms of reaching the desired object configuration. We also introduce a validity metric to evaluate whether the trajectory is reasonable for reaching the goal.  See Sec.~\ref{sec:exp_setup} for more details. 

%% file: method.tex
In this section, we describe our trajectory optimization formulation which considers the geometries of both the fingers and the object for contact and rolling constraints. The key questions are: (1) how to formulate a geometry-aware trajectory optimization problem that considers finger rolling and (2) how to process the geometries of both the robot and the object to formulate the contact and rolling constraints differentiably. Unlike previous methods~\cite{pang2023global, posa_contact_implicit}, our method accounts for the non-trivial geometry of robot fingers, enabling the formulation of finger rolling behavior to effectively tackle sensitive reorientation tasks, which demand precise manipulation and control.

In our method, the rolling behavior naturally emerges from the optimization. We do not manually specify any desired rolling behavior, allowing the algorithm the flexibility to choose whether and how to roll a finger during manipulation. 
\vspace{-0.7cm}
\subsection{Trajectory Optimization with 3D Finger Rolling}
\vspace{-0.1cm}
We focus on trajectory optimization with fixed contact modes. The robot fingers start the task in contact with the object. During manipulation, the robot can only utilize rolling to adjust the contact points locally but is not allowed to break and establish new contacts, such as through regrasping. We do not consider finger sliding. We also assume the system is quasi-static. 
The trajectory optimization is formulated as: 
\vspace{-0.2cm}
\begin{equation}
\begin{aligned}
& \min_{\substack{\mathbf{s}_1, \mathbf{s}_2, \cdots, \mathbf{s}_T; \\ \mathbf{u}_0, \mathbf{u}_1, \cdots, \mathbf{u}}_{T-1}}  J_{goal}(\bm{\tau}) + J_{smooth}(\bm{\tau}) \\
&\text{s.t.} \quad \mathbf{q}_{min} \leq \mathbf{q}_t \leq \mathbf{q}_{max}, \quad \mathbf{u}_{min} \leq \mathbf{u}_t \leq \mathbf{u}_{max} \\
&f_{contact}(\mathbf{s}_t) = 0, \quad f_{kinematics}(\mathbf{s}_t, \mathbf{s}_{t+1}) = 0 \\
&f_{balance}(\mathbf{s}_t, \mathbf{s}_{t+1}, \mathbf{u}_t) = 0\\
& f_{friction}(\mathbf{s}_t, \mathbf{u}_t) \leq 0, \quad f_{min\_f}(\bm{u}_t) \le 0 \\
\end{aligned}
\vspace{-0.1cm}
\end{equation}
where the state space is defined as $\mathbf{s}_t := \{\mathbf{q}_t, \mathbf{o}_t\}$. The action space is defined as $\mathbf{u}_t := \{\hat{\mathbf{u}}_t, \mathbf{f}_{1,t}, \cdots, \mathbf{f}_{N, t}, \mathbf{f}_{e,t} \}$, where $\hat{\mathbf{u}}_t$ is the delta action, 
$\mathbf{f}_{i, t}$ and $\mathbf{f}_{e,t}$ are the forces that the $i$th finger and the environment apply to the object respectively. The objective $J_{goal}$ incentivizes the robot to manipulate the object to the desired pose, 
and $J_{smooth}$ is defined as $J_{smooth} = \sum_{t=1}^T ||\mathbf{s}_t - \mathbf{s}_{t-1}||_2^2$. $f_{contact}$ ensures the fingers are always in contact with the object. $f_{kinematics}$, $f_{balance}$, and $f_{friction}$ are kinematics constraints, wrench balance constraints, and friction constraints, which are explained in the section below. $f_{min\_f}$ represents the constraint that ensures the force norm remains above a specified threshold, addressing the need to overcome joint friction and maintain contact in hardware experiment~\cite{murray2017mathematical}. Our primary contribution lies in the formulation of $f_{contact}$ and $f_{kinematics}$, utilizing SDF and a sampling-based method for estimating constraints and their gradients. 
We use CSVTO~\cite{power2024constrained} to solve the optimization. 
\subsubsection{Contact Constraints}
\label{sec:contact_constraint}
The constraints are formulated as $\Phi_i(\mathbf{q}_t, \mathbf{o}_t) = 0$,
where $\Phi_i$ returns the distance between the $i$th finger and the object. The distance between two shapes is defined as the minimal distance between any two points sampled from their surfaces. As robot fingers can have non-primitive geometries, estimating the constraints requires parameterizing the geometries. Additionally, the estimation must be differentiable to allow the gradients of the constraints to be used for updating the trajectory. We will further discuss the geometry processing in Sec.~\ref{sec:geom}.
\subsubsection{Kinematics Constraint}
\label{sec:kine_constraint}
As mentioned above, we do not consider finger slipping. Assuming pure rolling between the robot fingers and the object, the contact point on the finger and the contact point on the object have to share the same velocity. Mathematically, this can be written as: 
\vspace{-0.1cm}
\begin{equation}
    J_{\mathbf{c}_{i,t}}(\mathbf{q}_t) \dot{\mathbf{q}}_t = \bm{\omega}_t \times \mathbf{r}(\mathbf{c}_{i,t}, \mathbf{o}_t) + \dot{\mathbf{x}}_t,  
    \vspace{-0.1cm}
\end{equation}
where $J_{\mathbf{c}_{i,t}}$ is the Jacobian matrix at the contact point $\mathbf{c}_{i,t}$ on the robot, $\mathbf{c}_{i,t}$ denotes the contact point between the $i$th finger and the object at timestep $t$. $\mathbf{c}_{i,t}$ is a function of the state $\mathbf{s}_t$. $\dot{\mathbf{q}}_t$ is the joint velocity, approximated with finite difference: $\dot{\mathbf{q}}_t = (\mathbf{q}_{t+1} - \mathbf{q}_{t})/\Delta t$. $J_{\mathbf{c}_{i,t}}(\mathbf{q}_t)\dot{\mathbf{q}}_t$ is the velocity of the contact point $\mathbf{c}_{i,t}$.  $\bm{\omega}_t$ is the angular velocity of the object. $\mathbf{r}(\mathbf{c}_{i,t}, \mathbf{o}_t)$ outputs the radial vector from the object frame origin to the contact point $\mathbf{c}_{i,t}$. $\dot{\mathbf{x}}_t$ is the object's velocity. 

Though sharing the same contact velocities, the kinematics constraint only restricts the slipping behavior but allows finger rolling, as contact points might change and the kinematics constraint may apply to different contact points. 

However, having both the kinematics constraint and the contact constraint may over-constrain the problem: approximating velocities by finite differencing positions, both the contact constraint (1D constraint) and the kinematics constraint (3D constraint) specify the contact point location (3D variable), adding up to a 4D constraint. 
Thus, we project the 3D kinematics constraints into the 2D tangent space of the contacts: $R(\mathbf{n}_{i,t})(  J_{\mathbf{c}_{i,t}}(\mathbf{q}_t) \dot{\mathbf{q}}_t - \bm{\omega}_t \times \mathbf{r}(\mathbf{c}_{i,t}, \mathbf{o}_t) - \dot{\mathbf{x}}_t),$
where $R(\mathbf{n}_{i,t})$ is the projection matrix based on the contact normal $\mathbf{n}_{i,t}$ of the $i$th finger. 
See Fig.~\ref{fig:rolling_and_kinematics}b for visualization.

Similar to contact constraints, estimating $\mathbf{c}_{i,t}$, $J_{\mathbf{c}_{i,t}}$, $\mathbf{n}_{i,t}$ and their gradients also requires parameterizing the geometries, which is discussed in Sec.~\ref{sec:geom}.

\subsubsection{Wrench Balance Constraint}
With the quasi-static assumption, the system should satisfy wrench balance after applying actions. Specifically, at any timestep $t$, applying an action $\mathbf{u}_t$ to the state $\mathbf{s}_t$, the system changes its state to $\mathbf{s}_{t+1}$ and wrench balance should be satisfied with state $\mathbf{s}_{t+1}$. Wrench balance constraints naturally introduce a constraint involving $\mathbf{s}_t$, $\mathbf{s}_{t+1}$, and $\mathbf{u}_t$, which can be interpreted as the dynamics constraint of the system.

We assume the existence of a low-level PD controller. In quasi-static scenarios, only the proportional term contributes to the joint torque at $\mathbf{s}_{t+1}$. Therefore, wrench balance constraints can be written as: 
\vspace{-0.3cm}
{\allowdisplaybreaks
\begin{align}
    &K_p(\mathbf{q}_{t+1} - \mathbf{q}_t - \hat{\mathbf{u}}_t) - \bm{\tau}_{g}(\mathbf{q}_{t}) + \sum_i J^T_{\mathbf{c}}(\mathbf{q}_t)\mathbf{f}_{i,t}= 0 \label{equ:force_balance_robot}\\
    &\sum_{i}\mathbf{f}_{i,t} + \mathbf{f}_{e,t}=m\mathbf{g} \label{equ:force_balance_obj}
\end{align}
\vspace{0.1cm}
\begin{align}
    &\sum_i  \mathbf{f}_{i,t}\times\mathbf{r_i}(\mathbf{s}_{t})+ \mathbf{f}_{e,t}\times\mathbf{r_e}(\mathbf{s}_{t})=m\mathbf{r_{com}}(\bm{\theta}_{t+1}) \times \mathbf{g} \label{equ:torque_balance_object}
\end{align}
}
where $K_p$ is the gain matrix for the PD controller, $\bm{\tau}_{g}(\mathbf{q}_{t})$ is the joint torque required to counteract the gravity, and $\mathbf{g}= [0, 0, 9.8]^T$ is the gravity vector. $\mathbf{r_{i}}$, $\mathbf{r_{e}}$, and $\mathbf{r_{com}}$ are the vectors represented in the world frame pointing from the object frame origin to the contact point with the $i$th finger, the contact point with the environment and the object's CoM, respectively. $\mathbf{f}_{i,t}$ and $\mathbf{f}_{e,t}$ are the force that the $i$th finger and the environment apply to the object respectively. The visualization of the variables is shown in Fig.~\ref{fig:rolling_and_kinematics}c. Eq.~\ref{equ:force_balance_robot}, Eq.~\ref{equ:force_balance_obj} and Eq.~\ref{equ:torque_balance_object} refer to the torque balance for robot joints, the force and torque balance for the object respectively. 

Having both control input $\hat{\mathbf{u}}_t$ and contact forces $\mathbf{f}_{i,t}$ might seem redundant, since one can possibly derive $\mathbf{f}_{i,t}$ from $\mathbf{s}_t$ and $\hat{\mathbf{u}_t}$. However, explicitly computing the contact forces is not trivial. Previous work has proposed to include computing contact forces as part of the optimization problem~\cite{henze2016passivity, patel2019contact}. Similar to those methods, we add the forces $\mathbf{f}_{i,t}$ and $\mathbf{f}_{e,t}$ as decision variables and compute them within the optimization. The wrench balances also ensure the consistency of control actions $\hat{\mathbf{u}}_t$ and forces.
\subsubsection{Friction Constraints}
We use the Coulomb friction constraint:
$||\mathbf{f}_{i,t}^t|| \le \mu ||\mathbf{f}_{i,t}^n||,$
where $\mathbf{f}_{i,t}^t$ is the tangential force, $\mathbf{f}_{i,t}^n$ is the normal force, and $\mu$ is the friction coefficient. However, satisfying the second-order friction constraint can be numerically unstable when the tangential force is close to 0. To address this, we use a linearized 4-sided friction cone~\cite{ben2001polyhedral}, formulated as: 
$ A(\mathbf{n}_{i,t},\mu)\mathbf{f}_{i,t} \leq 0$,
where the matrix $A$ depends on the contact normal $\mathbf{n}_{i,t}$ and $\mu$.
\subsubsection{Trajectory Optimization}
The trajectory optimization problem is: 
\vspace{-0.4cm}
{
\allowdisplaybreaks
\begin{align}
& \min_{\substack{\mathbf{s}_1, \mathbf{s}_2, \cdots, \mathbf{s}_T; \\ \mathbf{u}_0, \mathbf{u}_1, \cdots, \mathbf{u}_{T-1}}}  J_{goal}(\bm{\tau}) + J_{smooth}(\bm{\tau}) \\
&\text{s.t.} \quad \mathbf{q}_{min} \leq \mathbf{q}_t \leq \mathbf{q}_{max}, \quad \mathbf{u}_{min} \leq \mathbf{u}_t \leq \mathbf{u}_{max}\\
&\Phi_i(\mathbf{q}_t, \mathbf{o}_t) = 0 \label{equ:final_contact}\\
&R(\mathbf{n}_{i,t})(J_{\mathbf{c}_{i,t}}(\mathbf{q}_t) \dot{\mathbf{q}}_t  - \bm{\omega}_t \times \mathbf{r}(\mathbf{c}_{i,t}, \mathbf{o}_t) - \dot{\mathbf{x}}_t) = 0\label{equ:final_kine}\\
&K_p(\mathbf{q}_{t+1} - \mathbf{q}_t - \hat{\mathbf{u}}_t) - \bm{\tau}_{g}(\mathbf{q}_{t}) + \sum_{i} J^T_{\mathbf{c}}(\mathbf{q}_t)\mathbf{f}_{i,t}= 0 \label{equ:final_torque_balance_robot}\\
&\sum_{i}\mathbf{f}_{i,t} + \mathbf{f}_{e,t}=m\mathbf{g} \label{equ:final_force_balance_obj}\\
&\sum_i  \mathbf{f}_{i,t}\times\mathbf{r_i}(\mathbf{s}_{t})+ \mathbf{f}_{e,t}\times\mathbf{r_e}(\mathbf{s}_{t})=m\mathbf{r_{com}}(\bm{\theta}_{t+1}) \times \mathbf{g}\label{equ:final_torque_balance_obj}\\
& A(\mathbf{n}_{i}(\mathbf{s}_t), \mu)\mathbf{f}_{i,t} \leq 0 \label{equ:friction}\\
& f_{i}^{min} - ||\mathbf{f}_{i,t}|| \leq 0 \label{equ:final_min_force} .  
\end{align} 
}
Eq.~\ref{equ:final_contact}, \ref{equ:final_kine}, and \ref{equ:friction} refer to $f_{contact}$, $f_{kinematics}$, and $f_{friction}$, and Eq.~\ref{equ:final_torque_balance_robot}, \ref{equ:final_force_balance_obj}, and \ref{equ:final_torque_balance_obj} refers to $f_{balance}$. Eq.~\ref{equ:final_min_force} refers to $f_{min\_f}$, with $f_i^{min}$ as the predefined force threshold.
\vspace{-0.2cm}
\subsection{Geometry Parametrization}
\label{sec:geom}
As described above, we utilize geometry information to formulate the constraints, such as contact constraints, and kinematics constraints. However, incorporating geometry information, such as meshes, as input is generally intractable. Computing geometry-related gradients for trajectory optimization algorithms adds further complexity. One of our main contributions can be summarized as using a sample-based method combined with \textit{softmin} to incorporate shape-related constraints and gradients.
Those include the distance function $ \Phi_i(\mathbf{q}_t, \mathbf{o}_t)$ for contact constraints, contact points $\mathbf{c}_{i,t}$, the jacobian matrix at the contact point $J_{\mathbf{c}_{i,t}}(\mathbf{q}_t)$, and the contact normal $\mathbf{n}_{i,t}$ for kinematics constraints. $\mathbf{n}_{i,t}$ is also used in formulating the friction constraints. 
In this section, we will ignore the timestep subscript $t$ for simplicity. 

We assume the manipulated object and the robot are described by primitive shapes and triangle meshes respectively.
Primitive shapes include boxes, cylinders, spheres, and their combinations. Many objects in real-world applications can be simplified as primitives or their combinations, e.g., a screwdriver as a combination of two cylinders with different radii. Primitives have an analytical formulation of their geometry and SDF, enabling efficient computation of contacts and distances. In the following text, we will utilize the function $\phi_k(\mathbf{p}_j, \mathbf{o})$ to query the SDF of the $k$th primitive of configuration $\mathbf{o}$ at position $\mathbf{p}_j$. The subscript $k$ is introduced because, in practice, the object can be a combination of multiple primitives. Thus, the distance between the point $\mathbf{p}_j$ and the object can be written as $\phi(\mathbf{p}_j, \mathbf{o}) = \min_k \phi_k(\mathbf{p}_j, \mathbf{o})$ 

\begin{figure}
    \vspace{0.2cm}
    \centering
    \includegraphics[width=0.75 \linewidth]{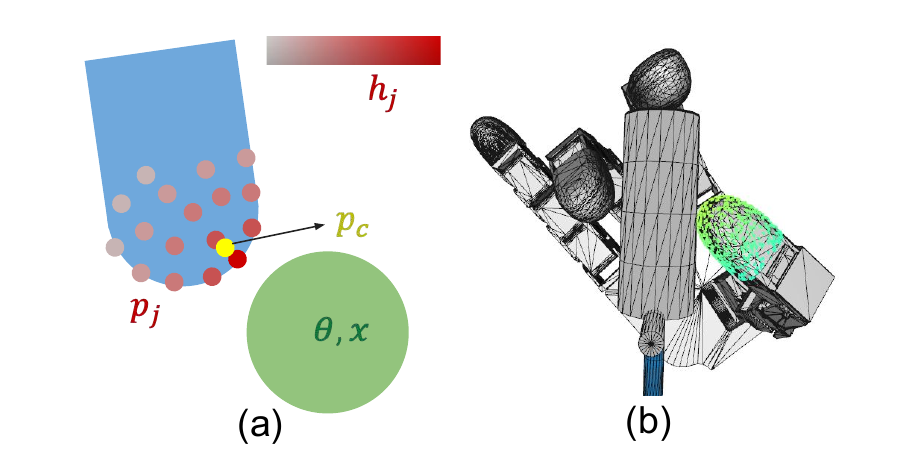}
    \vspace{-0.3cm}
    \caption{(a) Points (shown in red) are sampled over the surface of the robot. Points closer to the object are assigned a higher weight $h_j$, which is visualized with higher saturation. The contact point $\mathbf{p}_c$ is a weighted sum of $\mathbf{p}_j$. (b) Visualization of the sampled points on an actual robot}
    \vspace{-0.1cm}
    \label{fig:sample_points}
\end{figure}

However, not all objects can be simplified as primitives, e.g. the robot's fingertips. Given the meshes of non-primitive objects, we parameterize the mesh by uniformly sampling $N$ points $\mathbf{p}_j, j\in\{1,2,\cdots, N \}$ on the surface of the object, where $\mathbf{p}_j$ denotes the point coordinate in the world frame. See Fig.~\ref{fig:sample_points}. As for robot fingertips, points are first sampled in the corresponding robot link frame $\mathbf{p}_j^L$, and then we use forward kinematics to transform them into the world frame. Thus, $\mathbf{p}_j$ is a function of robot joint angles $\mathbf{q}$: $\mathbf{p}_j = f_p(\mathbf{p}_j^L, \mathbf{q})$. 
We will omit the function input and write it as $\mathbf{p}_j$ for simplicity and clarity. 


With the assumption of the primitive objects, we have access to the SDF of the object: $\phi(\mathbf{p}_j, \mathbf{o})$. The distance between the $i$th finger and the object can be written as: $\Phi_i(\mathbf{q}, \mathbf{o}) = \min_{j\in P_i} \phi(\mathbf{p}_j, \mathbf{o})$, 
where $P_i$ is the set of points sampled on the surface of the $i$th fingertip. While this provides an accurate estimation with enough sampled points, the minimization operation is not differentiable and also discards information from the other points. Gradients from these points are important for finger rolling as they convey details about the shape of the object, containing information about how to roll the fingers on the surface of the manipulated object based on their geometries. This information becomes critical when precise control is necessary to successfully complete the task.
To address the differentiablity problem, we use the \textit{softmin} to produce a weighted summation over all the points. The weight $h_j$ for the $j$th point is given by:
\begin{equation}
    h_j = \frac{\exp(-\delta \phi(\mathbf{p}_j, \mathbf{o}))}{\sum_k \exp(-\delta \phi(\mathbf{p}_k, \mathbf{o}))},
    \vspace{-0.2cm}
\end{equation}
where $\delta$ is the temperature. In practice, we use a high temperature to make the \textit{softmin} approximation closely match the actual distance. The estimated distance is written as:
\begin{equation}
    \Phi_i(\mathbf{q}, \mathbf{o}) = \sum_{j\in P_i} h_j \phi(\mathbf{p}_j, \mathbf{o}).
    \vspace{-0.2cm}
\end{equation}
Similarly, we also use \textit{softmin} to estimate the closest points (contact points) $\mathbf{c}_i$ between the finger and the object, and the Jacobian matrix $J_{\mathbf{c}_i}(\mathbf{q})$ at the contact point: 
\begin{equation}
    \mathbf{c}_i = \sum_{j\in P_i} h_j \mathbf{p}_j, \quad J_{\mathbf{c}_i}(\mathbf{q}) = \sum_{j\in P_i} h_j J_{\mathbf{p}_j}(\mathbf{q}).
    \vspace{-0.2cm}
\end{equation}
The contact normal $\mathbf{n}_i(\mathbf{p}, \mathbf{o})$ is estimated as the gradient of $\phi(\mathbf{p}_j, \mathbf{o})$ w.r.t. $\mathbf{p}_j$ using \textit{softmin}, ensuring differentiability:
\begin{equation}
    \mathbf{n}_i(\mathbf{p}, \mathbf{o}) = \sum_{j\in P_i} h_j \frac{\partial \phi(\mathbf{p}_j, \mathbf{o})}{\partial \mathbf{p}_j}
\end{equation}

As for the gradients of the constraints mentioned above, the trajectory optimization only takes in the gradients w.r.t. $\mathbf{s}$ and $\mathbf{u}$. While packages that use autograd  (e.g. PyTorch) can compute these gradients, they can be quite slow for a time-sensitive MPC framework. To address this, we derive analytical forms as much as possible to speed up the computation. Specifically, we apply the chain rule to get the analytical formulations of the gradients of distance queries $\Phi_i(\mathbf{q}_t, \mathbf{o}_t)$, contact points $\mathbf{c}_{i,t}$, the jacobian matrix $J_c(\mathbf{q}_t)$, and the contact normal $\mathbf{n}_{i,t}$ w.r.t. $\mathbf{s}_t$ and $\mathbf{u}_t$. We use PyTorch to compute the rest of the gradients.

%% file: experiments.tex
In this section, we aim to verify: (1) whether finger-rolling behavior is achievable from the optimization, (2) whether finger-rolling and geometry information improve the performance over baseline, (3) whether our method works on real-world systems, and (4) whether our method is fast enough to be applied in an MPC framework.

To answer the questions above, we propose a benchmark for multi-finger dexterous manipulation including five tasks that vary in difficulty. We also test our method and one of the best-performing baselines in a real-world setup. 

\subsection{Benchmark Setup}
\label{sec:exp_setup}
\begin{figure}[htbp]
    \centering
    \includegraphics[width=0.485\textwidth]{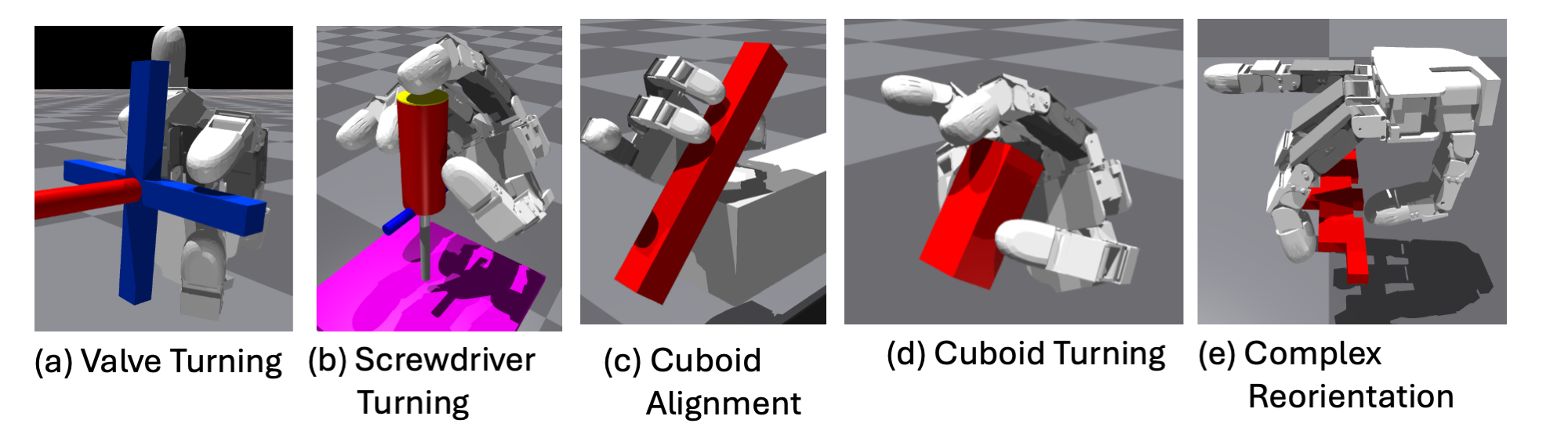}
    \vspace{-0.6cm}
    \caption{The benchmark consists of multi-finger manipulation tasks with and without extrinsic environment contacts.}
    \label{fig:benchmark}
\end{figure}

The task setup for each experiment is shown in Fig.~\ref{fig:benchmark}. The Allegro hand~\cite{allegro}, which consists of four fingers and sixteen servo-driven joints, is used for all the tasks. We only use the thumb, index finger, and middle finger, as three fingers are sufficient for all the tasks we consider. Additionally, we exclude wrist movements, relying solely on finger action. Isaac Gym is used as our simulator~\cite {makoviychuk2021isaac}.

\textbf{Valve Turning(VT):} The object is a cross-shaped valve with one revolute joint, consisting of two cuboids. The goal is to turn the valve by $45^\circ$, set relatively close as the robot cannot rotate the valve significantly without changing the contact mode.

\textbf{Screwdriver Turning(ST):} 
We assume the screwdriver is already mated with the screw head and focus on turning. The goal is to turn the screwdriver 90 degrees while keeping it upright. The screwdriver turning task can benefit from rich finger-rolling behavior. It also tests the performance of our method on everyday tool manipulation tasks. 

Simulating the screwdriver turning behavior is challenging due to the complex screw thread geometry and the interaction between the screwdriver tip and the screw. To simplify, we approximate these interactions in the simulator by attaching the screwdriver to a table using a spherical joint, assuming it remains seated in the screw head. We apply damping in the yaw direction to approximate the required turning torque. We model the screwdriver as two cylinders. In hardware experiments, we use an actual screw.

We use a precision screwdriver, usually used in repairing electronics, which has a revolute joint on the top of the body. Humans usually push the top with their index finger and turn it with the thumb and middle finger. 
This kind of screwdriver and its specific way of turning it can highlight the importance of dexterous manipulation and finger rolling. 

\textbf{Cuboid Alignment(CA):} Peg-in-hole tasks are important in industrial applications such as robot assembly. Usually, the first step to insert a peg is to reorient and align it with the hole. Reorienting the peg requires finger rolling. In this task, we model the peg as a cuboid. Additionally, we include a box to aid with completing the task, e.g., to reorient the cuboid to an upright pose, one can push the cuboid to the vertical surface of the box. 
This experiment also highlights the versatility of our framework, because including extrinsic contact only requires adding one constraint.

The environment consists of a cuboid(peg) and a box. The cuboid starts in contact with the box. The robot needs to utilize the box to turn the cuboid for approximately 45 degrees to make it upright.

To encourage contact between the peg and the box, we add an additional contact constraint to our trajectory optimization: 
$\Phi_{peg}(\mathbf{o}_t) = 0,$
where $\Phi_{peg}$ returns the distance between the peg and the box. 

\textbf{Cuboid Turning(CT):} 6D in-hand reorientation is a fundamental task in multi-finger manipulation. We first start with reorienting a cuboid. The robot starts with the thumb and middle finger positioned on the opposite sides of the cuboid to ensure stability. The objective is to turn the cuboid by $60^\circ$, reorienting it horizontally. Unlike most RL methods focusing on in-hand reorientation~\cite{qi2023hand, huang2021generalization}, the hand in our setup faces downward instead of upward, making dropping the object more likely.

\textbf{Complex Reorientation(CR): }This task also focuses on 6D in-hand reorientation but is more challenging than the cuboid turning task. The robot needs to reorient a thin and complex object composed of multiple cuboid primitives. 
This task is designed to evaluate the ability to manipulate complex objects composed of multiple primitives.

In our experiments, we focus exclusively on the stage where the robot has already established contact with the object. Simulations are initialized either with the robot in contact or using a pregrasp algorithm, detailed in Appendix.

For the trajectory optimization problem setup, we use the horizon $T=12$ steps for ST and $T=10$ steps for the rest of the tasks, based on how much the object needs to rotate. The horizon $T$ is reduced by 1 after each step of execution. 
To solve the trajectory optimization, we use relatively long iterations at the first step to warm up, then much shorter iterations for the remaining steps in the simulation.
The solution of the previous step is used to initialize the solution for the current step. The initialization for robot states and actions at the first step is sampled from normal distributions. See details in the appendix. 
Object configurations $\mathbf{o}_t$ are initialized via linear interpolation between the current and goal configurations.
\begin{table*}
    \vspace{0.2cm}
    \centering
    \begin{tabular}{|c|c|c|c|c|c|c|c|c|c|c|}
    \hline
    \multirow{2}{*}{ Method} & \multicolumn{2}{c|}{Valve Turning}& \multicolumn{2}{c|}{Screwdriver Turning}& \multicolumn{2}{c|}{Cuboid Alignment} &\multicolumn{2}{c|}{Cuboid Turning} & \multicolumn{2}{c|}{Complex Reorientation} \\
    \cline{2-11}
    & distance & valid & distance &valid & distance &valid& distance &valid& distance &valid\\
    \hline
    Ours & $5.45^\circ \pm 2.67^\circ$ & n.d. & $22.84^\circ\pm5.61^\circ$ & \textbf{0.93} & $14.84^\circ\pm1.05^\circ$ & \textbf{1.0} & \bm{$4.57^\circ\pm1.02^\circ$} & \textbf{1.0} & \bm{$21.78^\circ\pm3.59^\circ$}&0.97\\
    \hline
    Ablation & $3.45^\circ \pm 1.93^\circ$ & n.d. & \bm{$21.16^\circ \pm 4.18^\circ$} & 0.07 & $18.66^\circ \pm 4.43^\circ$ & 0.83 & $72.12^\circ \pm 17.21^\circ$ & 0.23 & $35.04^\circ \pm 20.11^\circ$ & 0.47\\
    \hline
    PtC & $8.02^\circ \pm 3.97^\circ$ & n.d. & No valid trial & 0.0 & No valid trial & 0.0 & $8.01^\circ \pm 3.97^\circ$ & \textbf{1.0} & No valid trial & 0.0 \\
    \hline
    MPPI &$1.77^\circ \pm 1.74^\circ$ & n.d. & $33.32^\circ \pm 14.61^\circ$& 0.83 & \bm{$5.67^\circ \pm 1.75^\circ$} & 0.87 & $8.65^\circ \pm 6.26^\circ$ & 0.80 & $28.90^\circ \pm 26.78^\circ$ & 0.53 \\
    \hline
    RL & \bm{$0.65^\circ \pm 0.00^\circ$} & n.d. & No valid trials & 0.0 & $90.01^\circ \pm 0.02^\circ$ & \textbf{1.0} & $90.39^\circ \pm 1.21^\circ$ & \textbf{1.0} & $58.98^\circ \pm 34.03^\circ$ & \textbf{1.0} \\
    \hline
    \end{tabular}
    \vspace{-0.1cm}
    \caption{Results are evaluated for 30 episodes in the simulation. N.D. means not defined.}
    \label{tab:simulation_result}
    \vspace{-0.8cm}
\end{table*}
In the simulation experiments, we consider two metrics: 

\textbf{Validity: } The algorithm may exploit simulator artifacts to achieve the goal. To address this, we design a validity metric based on human task performance to capture physical plausibility. Specifically, the screwdriver should be upright. A trajectory is considered valid if the center of the top of the screwdriver handle remains within 2cm of its starting position. For the remaining 6D reorientation tasks, validity is defined as not dropping the object. validity is not applicable in the valve turning tasks (marked as n.d. in Tab.~\ref{tab:simulation_result}).

\textbf{Distance to Goal: }
This metric evaluates the distance between the object's final and desired orientation, defined as the relative angle between two SO(3) orientations. Only valid trajectories are considered for the evaluation.

\subsection{Baselines and Ablations}
We compare our method with the following 4 methods: 

\textbf{Planning to Contact (PtC):} The robot starts in contact with the object. Given the initial contacts, the fingertip (end effector) positions $\hat{\mathbf{p}}_i$ in the object frame are recorded. By linearly interpolating the object pose between the current and goal poses, we get the desired object pose for each time step. Assuming the fingertip positions do not change in the object frame, we determine the desired fingertip locations at each time step in the world frame. An inverse kinematics (IK) problem is then solved to compute the robot's joint angles required to reach these positions. 
This method does not consider rolling or object geometry. The idea of searching the end effector pose first and then solving an IK is used in many robot manipulation tasks, especially in the motion planning domain~\cite{vahrenkamp2009humanoid, cohen2010search, ahuactzin1999kinematic}.

\textbf{Model Predictive Path Integral (MPPI): }MPPI is a popular model-based trajectory optimization method. 
We implement MPPI as proposed in~\cite{williams2017information}, using the simulator as the dynamics model to roll our trajectories, similar to~\cite{pezzato2023sampling}. Since the simulator requires both configurations and velocities as the state, we provide MPPI with additional velocity information, whereas other methods only utilize configurations.
For valve turning, we use 50 particles for sampling, i.e., at each time step, we spawn 50 copies of the environment, each executing different action sequences sampled. The sampling horizon is set to be 4. We experimented with extending the horizon, but the performance showed no significant improvement. For the remaining tasks, we use 500 particles. Additionally, we set the sampling horizon $T = 2$. We attempted setting $T$ the same as that in our method, but the robot almost always dropped the object due to the sensitive nature of the tasks: it is nearly impossible to constantly satisfy the stringent contact constraints in a lower-dimensional manifold over an extended horizon through random action sampling, resulting in dropping the object frequently and high costs for almost every trajectory. Longer horizons generally result in poorer performance in our experiments. Since MPPI performs rollouts and evaluates within the exact same environment, it should, in theory, reach the goal for all tasks given enough sampling budget and an appropriate cost function. To ensure a fair comparison, we match the sampling budget to the runtime of our method.

\textbf{Reinforcement Learning:} Although RL has a different problem setup , requiring training on the specific task prior to evaluation, we include it because it is a popular approach for multi-finger manipulation. We use HORA~\cite{qi2023hand}, as it also addressed in-hand reorientation, though with the hand facing upwards. While HORA introduces a method for sim-to-real transfer, we evaluate it solely in simulation experiments, eliminating the need for transfer. For each task, we train three different seeds and train until it has fully converged.

\textbf{Ablation without Geometry-Based Constraints: } To demonstrate the importance of differentiable geometry information and rolling constraints, we remove the contact constraints (Eq.~\ref{equ:final_contact}) and the kinematics constraints (Eq.~\ref{equ:final_kine}). Instead, we replace those constraints with simplified contact constraints. 
Similar to the PtC method, we assume the fingertip locations in the object frame do not change:
$f_{FK,i}(\mathbf{q}_t) - f_{trans}(\mathbf{o}_t, \hat{\mathbf{p}}_i) = 0,$
where $f_{FK,i}$ is the forward kinematics function to get the fingertip location for the $i$th finger, $f_{trans}$ returns the desired fingertip location in the world frame based on the current object pose $\mathbf{o}_t$. Unlike PtC, this method considers friction, joint limits, and force balances. Additionally, the optimizer selects the next object pose rather than linear interpolation.
\subsection{Simulation experiments}
We ran 30 trials for each method on each task. The results are shown in Table~\ref{tab:simulation_result}. Our method is the only one to achieve validity rates close to 1 across all tasks while coming close to the desired orientation. Rolling behaviors are also observable in our experiments (see video). Our method does sometimes produce invalid trajectories. This is due to either an insufficient optimization budget or constraint violation between time steps. Specifically, our method only considers constraint at the discretized time steps. However, these constraints may be violated when the robot executes actions and transitions between neighboring time steps.

For the PtC method, the desired end effector positions are usually unreachable, causing the IK solver to fail frequently. For example, in the screwdriver turning task, the thumb is almost fully extended and has a limited configuration space, which requires rolling the thumb to turn the screwdriver further. Without a valid IK solution, the robot's behavior becomes unpredictable, leading to dropping the object frequently. However, it achieves a validity rate of $1.0$ on the cuboid turning task, as the thumb and middle require only minimal movement and these two fingers provide additional stability to prevent object dropping.

The MPPI method is very effective in exploring the desired turning behavior. It can achieve a similar distance to the goal as our method, though with a slightly lower validity rate. It even has slightly better performance on the valve turning and cuboid alignment task, as it does not need to adhere to the assumption of a fixed contact mode and quasi-static execution. However, MPPI does not consider the robustness explicitly and tends to end up with sensitive trajectories. When executing MPPI in the real-world setup, since the dynamics of the simulator no longer match the evaluation environment, i.e., the real-world setup, the performance is much worse. We will discuss it more in Sec.~\ref{sec:real_world}.

In our experiment, the RL method faces significant challenges with exploration: It is heavily penalized for dropping the object in the reward function, and it's hard to sample a long sequence of actions that adheres to the stringent contact required to prevent object dropping. As a result, it often converges to a conservative policy. For instance, in the cuboid alignment task, the RL method places the cuboid on the gray box to avoid dropping it, completely abandoning the objective of reorienting to the goal. However, we do not claim that RL methods are inherently unsuitable for our benchmark. With appropriate reward engineering and curriculum learning, it is possible to train an RL for in-hand reorientation with the hand facing downwards~\cite{chen2022system}. Our goal is to highlight the non-triviality of training a successful RL agent for our benchmark.

The ablation method, which excludes geometry-based constraints, still considers the rest of the constraints to prevent dropping the object. 
However, simply tracking the desired fingertip location can lead to unmodeled rolling behavior, pushing the object away from the goal, as demonstrated in the screwdriver task, or preventing the desired rolling behavior from occurring as demonstrated in the cuboid alignment task. In general, tasks that inherently require more finger rolling exhibit a larger performance gap between the ablation and our method, highlighting the importance of modeling rolling in dexterous manipulation.
\begin{table}
    \centering
    \begin{tabular}{|c|c|c|c|c|c|c|c|c|}
    \hline
          \multirow{2}{*}{ Task} & \multicolumn{2}{c|}{Ours} & \multicolumn{2}{c|}{Ablation} & PtC & \multicolumn{2}{c|}{MPPI} & RL \\
         \cline{2-9}
          & w.u. & online& w.u. & online & online & w.u. & online & online \\ 
         \hline
         VT& 12.9 s & 3.9 s& 17.8 s& 5.3 s& 8.7 s& 13.5 s& 4.5 s& $5 \times 10^{-4}$ s\\
         \hline 
         ST & 66.5 s& 10.0 s& 65.8 s& 9.8 s& 28.3 s& 78.0 s& 16.4 s& $6 \times 10^{-4}$ s\\
         \hline
         CA & 64.9 s& 10.1 s& 70.5 s&11.0 s& 46.4 s& 67.0 s& 14.4 s& $5 \times 10^{-4}$ s\\
         \hline
         CT & 39.9 s& 5.6 s& 46.5 s& 7.0 s& 8.7 s& 66.1 s& 13.3 s& $4 \times 10^{-4}$ s\\
         \hline
         CR & 39.8 s& 5.8 s& 43.4 s& 6.4 s& 50.7 s& 54.3 s& 28.9 s& $5 \times 10^{-4}$ s\\
         \hline
    \end{tabular}
    \vspace{-0.1cm}
    \caption{Computation time for the experiments. w.u. means warm up.}
    \label{tab:computation_time}
\end{table}
The computation time is shown in Tab.~\ref{tab:computation_time}.
For PtC, long iterations for solving IK result in extended runtimes when no valid solution is found.
Clearly, none of these methods except RL have adequate computation time for real-time MPC.
We did not focus on computational efficiency for this work and all methods were implemented in Python. We expect that optimizing the code and implementing our approach in C++ will lead to significant speed improvements. In the hardware experiment in Sec.~\ref{sec:real_world}, we reduce online iterations to speed up execution with minimal impact on performance.
\vspace{-0.3cm}
\subsection{Real-World Experiments}
\vspace{-0.2cm}
\label{sec:real_world}
We consider the screwdriver turning and cuboid alignment for real-world experiments. For screwdriver turning, we place a screw in a loose slot as the Allegro hand cannot exert large torques for tightening the screw. Our screwdriver is 3D-printed because the Allegro hand is larger than a human hand, making a standard screwdriver unsuitable for this robot. Similarly to a regular screwdriver, a high-friction tape is attached to the screwdriver body. The high-friction tape is also used in the cuboid alignment task. We use Aruco tags~\cite{garrido2014automatic} to estimate the poses of the screwdriver, the cuboid, and the robot hand (see Fig.~\ref{fig:pull_figure}). The PID values are manually adjusted to closely approximate a critically damped response.
\begin{table}
    \centering
    \begin{tabular}{|c|c|c|c|}
        \hline
        Task & Method& Distance to Goal & Validity \\
        \hline
         \multirow{2}{*}{ Screwdriver Turning} & Ours & $25.02^\circ \pm 6.66^\circ$ & 0.9\\
         \cline{2-4}
         & MPPI & No valid trial & 0.0 \\
         \hline
         \multirow{2}{*}{Cuboid Alignment}& Ours & $13.28^\circ \pm 1.35 ^ \circ$ & 0.9 \\
         \cline{2-4}
         & MPPI & No valid trial & 0.0 \\
         \hline
    \end{tabular}
    \vspace{-0.1cm}
    \caption{Hardware experiment results for 10 trials}
    \vspace{-0.3cm}
    \label{tab:hardware}
\end{table}
We compare our method with MPPI, the best-performing baseline over 10 trials. The results are shown in Tab.~\ref{tab:hardware}. For our method, the robot drops the object once in both tasks. The average distance to the goal is very similar to the results in the simulation. 
MPPI plans trajectories without considering constraints, making these trajectories sensitive to errors. As a result, perception noise and the dynamics mismatch between simulation and the real world lead to MPPI having no valid trials.

The sim2real gap can be attributed to: (1) Perception errors: inaccurate object pose estimation, and (2) imperfect joint modeling: unmodeled factors such as backlash and joint friction. Hardware experiments show our method is effective in simulation and transferable to the real world, though perception and modeling challenges remain.

%% file: discussion.tex
Experiments on our proposed benchmark suggest that sensitive multi-finger manipulation tasks benefit significantly from differentiable contacts and finger-rolling formulations. By adopting a sampling method to approximate the finger geometries and integrating it with the object SDF, we estimate the contact-related variables differentiably. The formulations enable a gradient-based trajectory optimizer to handle non-trivial geometries and extend the capabilities of dexterous manipulation. 
Our future work will integrate contact mode planning to allow changing finger contacts during tasks.
